\documentclass[letterpaper, 10 pt, conference]{ieeeconf}  % Comment this line out if you need a4paper

\IEEEoverridecommandlockouts                              % This command is only needed if 
\title{\LARGE \bf
CooperRisk: A Driving Risk Quantification Pipeline
\\ with Multi-Agent Cooperative Perception and Prediction
}

\author{Mingyue Lei$^{1,2*}$, Zewei Zhou$^{1*}$, Hongchen Li$^{2}$, Jia Hu$^{2\dagger}$ and Jiaqi Ma$^{1\dagger}$% <-this % stops a space
\thanks{$^{*}$Equally contributed to the work.}
\thanks{$^{\dagger}$Corresponding author.}
\thanks{$^{1}$Mingyue Lei, Zewei Zhou and Jiaqi Ma are with UCLA Mobility Lab, University of California, Los Angeles, USA. This work was done during Mingyue Lei's visit at UCLA Mobility Lab.
        {\tt\small mingyue2024@g.ucla.edu, zeweizhou@g.ucla.edu, jiaqima@g.ucla.edu}}%
\thanks{$^{2}$Mingyue Lei, Hongchen Li and Jia Hu are with the Key Laboratory of Road and Traffic Engineering of the Ministry of Education, Tongji University, Shanghai, China. 
        {\tt\small mingyue\_l@tongji.edu.cn, 2410195@tongji.edu.cn, hujia@tongji.edu.cn}}%
}

\usepackage{amsmath}
\usepackage{amsfonts}
\usepackage{multirow}
\usepackage{graphicx}
\usepackage{booktabs}
\usepackage{wrapfig}
\usepackage{tabularray}
\usepackage[capitalize]{cleveref}
\usepackage{subcaption}
\usepackage[compress]{cite}
\usepackage[table]{xcolor}

\begin{document}

\maketitle
\thispagestyle{empty}
\pagestyle{empty}

%%%%%%%%%%%%%%%%%%%%%%%%%%%%%%%%%%%%%%%%%%%%%%%%%%%%%%%%%%%%%%%%%%%%%%%%%%%%%%%%
\begin{abstract}

Risk quantification is a critical component of safe autonomous driving, however, constrained by the limited perception range and occlusion of single-vehicle systems in complex and dense scenarios. Vehicle-to-everything (V2X) paradigm has been a promising solution to sharing complementary perception information, nevertheless, how to ensure the risk interpretability while understanding multi-agent interaction with V2X remains an open question. In this paper, we introduce the first V2X-enabled risk quantification pipeline, \textit{CooperRisk}, to fuse perception information from multiple agents and quantify the scenario driving risk in future multiple timestamps. The risk is represented as a scenario risk map to ensure interpretability based on risk severity and exposure, and the multi-agent interaction is captured by the learning-based cooperative prediction model. We carefully design a risk-oriented transformer-based prediction model with multi-modality and multi-agent considerations. It aims to ensure scene-consistent future behaviors of multiple agents and avoid conflicting predictions that could lead to overly conservative risk quantification and cause the ego vehicle to become overly hesitant to drive. Then, the temporal risk maps could serve to guide a model predictive control planner. We evaluate the \textit{CooperRisk} pipeline in a real-world V2X dataset V2XPnP, and the experiments demonstrate its superior performance in risk quantification, showing a 44.35\% decrease in conflict rate between the ego vehicle and background traffic participants.

\end{abstract}

%%%%%%%%%%%%%%%%%%%%%%%%%%%%%%%%%%%%%%%%%%%%%%%%%%%%%%%%%%%%%%%%%%%%%%%%%%%%%%%%
\section{INTRODUCTION}
\label{sec:intro}

% Safety is the foremost concern in autonomous driving. Understanding the dynamic driving environment and quantifying the safety risk is the foundation of decision-making and planning in autonomous driving.

Connected autonomous vehicle (CAV) technology has been regarded as one of the most promising solutions for safe transportation \cite{han2024foundation, hu2023planning, zhou2022comprehensive}. To ensure safety, CAVs should accurately perceive their surroundings, quantify the driving risks, and navigate safely through dynamic environments. Despite the advancements in risk quantification (riskQ) \cite{yang2024safe, katrakazas2019new}, existing work still relies on single-vehicle information, and overlooks the potential risks posed by the objects in occlusion and outside the perception range \cite{wang2022potential}. Inherent perception errors are also ignored in the independent risk quantification modules, leading to unreal results. The community has witnessed the development of cooperative perception \cite{xiang2024v2x, yangjie2024towards}, where multiple agents (vehicles or infrastructures) share complementary information with Vehicle-to-Everything (V2X) communication. However, how to accurately quantify the driving risks with V2X information remains underexplored, especially for the systems that cover the entire pipeline from perception to planning.

\begin{figure}[t]
  \centering
  \includegraphics[width=\linewidth]{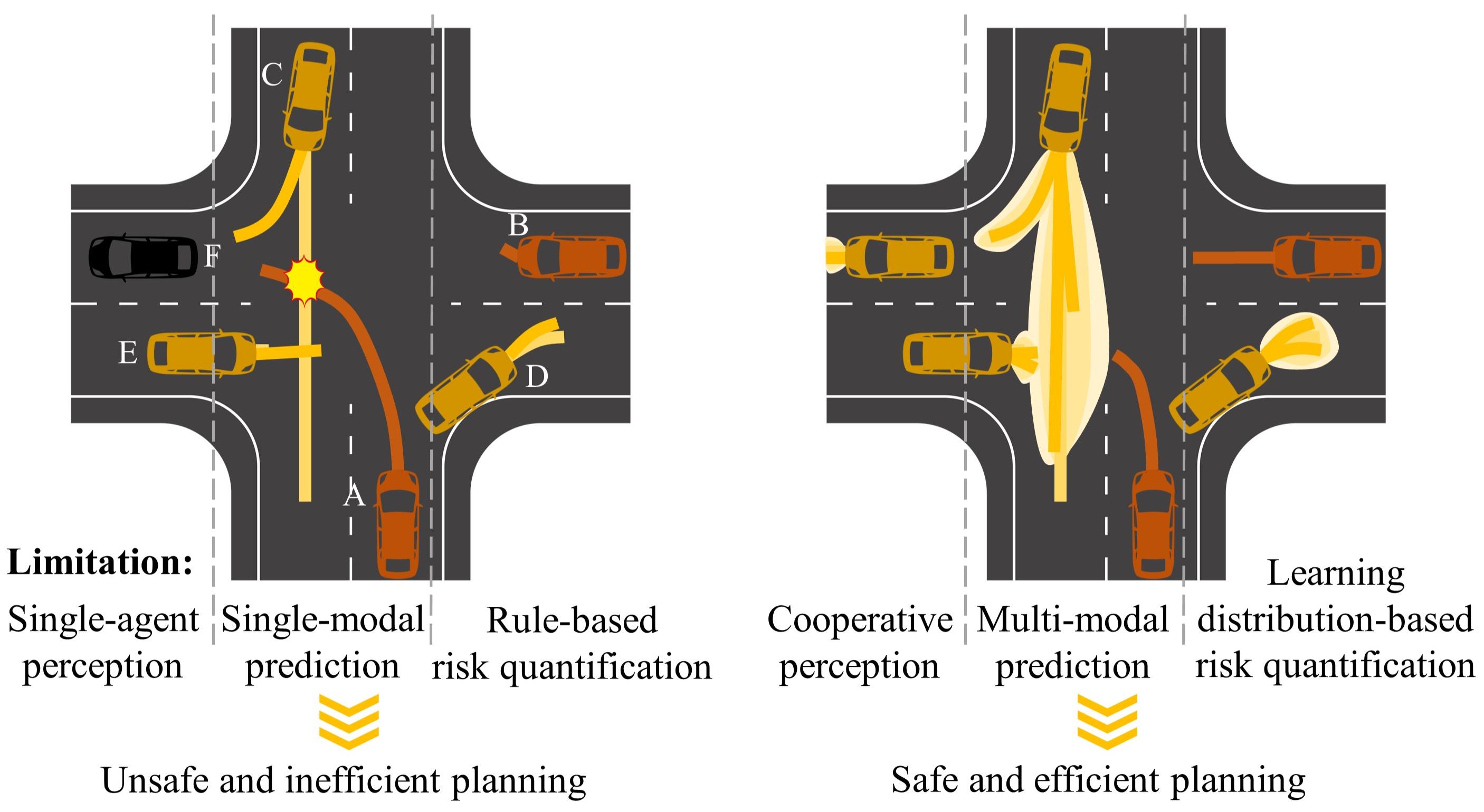}
  \caption{Unsafe planning may result from the limitations of single-agent perception, single-modal prediction and rule-based risk quantification. Detailed explanation is presented in the \cref{sec:intro}.} 
  \label{Figure1}
  \vspace{-0.7cm}
\end{figure}

In the field of risk quantification, rule-based and learning-based methods are two typical approaches. The former are interpretable but fail to cover the complex interaction, especially in dense scenarios, \cite{koopman2019autonomous, nadimi2020evaluation}, which encourages the community to explore the learning-based method. Due to the lack of the driving-risk ground truth, reinforcement learning (RL) is utilized to learn the risk metric \cite{yang2024safe}. Nevertheless, RL reasoning is poor in interpretability and its reward function is difficult to determine and still based on human expertise. To ensure risk interpretability while understanding multi-agent interaction, we adopt the future trajectories as the learning target to capture the interactions between objects and generate a scenario risk map based on the predicted trajectory distributions. Our risk map is defined as a set of maps at future timestamps, and each one shows the potential ego risk that varies with positions and modalities. Our risk maps are generated from the objects' future behaviors, highlighting the severity and exposure distributions of the risk to the ego CAV, rather than solely relying on the current information in previous works. This strategy aligns more closely with intuitive driver behavior in risk avoidance, enhancing the effectiveness of downstream risk-aware planning tasks.

This integrated pipeline of prediction and risk quantification has been preliminarily explored in \cite{shao2024uncertainty, zhang2022integrated}, however, these prediction models operate the prediction one object by one object, leading to unreasonable overlap of prediction results among multiple agents. The simple single-agent prediction always misleads the risk quantification module and narrows the drivable areas for the ego CAV. Hence, risk-oriented prediction should involve multiple agents and multiple modalities. Multi-modal prediction can differentiate objects' various behaviors, while multi-agent prediction ensures compatibility among different object behaviors, minimizing their self-collisions \cite{chen2022scept}. In detail, we adopted a unified transformer to encode history trajectories and multi-agent interaction. Then, a multi-intention-aware decoder is employed to simultaneously predict the multi-modal future motions of all objects within the scenario.

% Additionally, the multi-agent framework naturally aligns with multi-modality, as multi-agent prediction requires matching each agent’s intention by selecting suitable modes from the candidate multi-modal options.
In conclusion, we present \textbf{\textit{CooperRisk}}, a cooperative perception-prediction-riskQ pipeline, to comprehensively quantify risk in complex scenarios with V2X information. Fig. \ref{Figure1} shows an example to conclude the advantages of our \textit{CooperRisk}. The figure on the left shows the limitations in traditional systems: \textit{(1) Single-agent perception} prevents CAV A from being aware of vehicle F in its target lane. \textit{(2) Single-modal prediction} results in vehicle C's future trajectory being a right turn (dark yellow), while the ground truth is a straight one (light yellow), leading to the collision planning of vehicle A with vehicle C. \textit{(3)  Rule-based risk quantification} causes vehicle B to decelerate abruptly in response to vehicle D, which is making a right turn and will not enter B’s lane. This misjudgment arises because the risk quantification method considers only trajectory proximity, ignoring scene-consistent risk exposure and severity. In contrast, our \textit{CooperRisk} framework adopts \textit{(i) Cooperative perception} to ensure comprehensive awareness of the scenario, \textit{(ii) Multi-agent multi-modal prediction} to mitigate inter-trajectory collisions and differentiate different potential risk modalities, accurately delineating drivable areas for the ego CAV, \textit{(iii) Learning distribution-based risk quantification} to determine risks in a manner consistent with human drivers' evaluations of risk exposure and severity. Furthermore, the proposed riskQ method combines both rule-based and learning-based methods to ensure interpretability and accuracy. Our contributions can be summarized as follows: 

% For trajectory prediction model, it has drawn the community’s attention \cite{huang2022survey}, and the features of interaction, map, and history information have been considered in the emerging model. Compared to traditional single-vehicle trajectory prediction, risk-oriented trajectory prediction involves multi-agent multi-modal prediction. Multi-modal prediction can differentiate between various potential risks in the same scenario, while multi-agent prediction ensures compatibility among different object behaviors, minimizing self-collisions in prediction trajectories. 

% Then, we construct a risk map at each future time step based on the objects' positions, headings, and velocities to quantify the risks in the future scenario.

% To address those gaps, we present CooperRisk, a cooperative perception-prediction-riskQ pipeline, to make comprehensive risk quantification in complex scenarios with V2X information. Moreover, its prediction-risk part is a combination of the rule-based and learning-based model to ensure interpretability and accuracy of risk quantification. Furthermore, the prediction model is specifically designed for scenario risk incorporating both multi-modal and multi-agent prediction settings to differentiate different risks and ensure the subsequent planning will not be too afraid to drive with the accumulation of various risks. 
% Our contributions can be summarized as follows: 

\begin{itemize}
\item We propose a cooperative perception-prediction-riskQ pipeline, \textit{CooperRisk}. To the best of our knowledge, this is the first driving risk quantification pipeline enhanced by V2X information, which covers the entire process from perception to planning.
\item We formulate the risk quantification task as two stages to improve the interpretability and accuracy: risk-oriented prediction and risk map generation. The risk map provides the dynamic severity and exposure of risk over future frames to the ego CAV.
\item We specifically design a risk-oriented prediction, involving multi-modal multi-agent prediction to ensure scene consistency and accurately delineate drivable areas.
\item We evaluate the proposed \textit{CooperRisk} pipeline on a real-world V2X dataset V2XPnP, and validate it with a model predictive control planner, which demonstrates the superior performance in risk-aware planning.

\end{itemize}

\section{Related Work}

\noindent \textbf{Driving Risk Quantification.} Rule-based and learning-based methods are two typical approaches for quantifying risk \cite{lefevre2014survey}. Rule-based methods evaluate the risk with a series of hand-crafted rules, such as artificial potential field-based method \cite{li2020risk} and limiting surrogate measurements (time-to-collision \cite{eggert2014predictive} and time-to-reaction \cite{wagner2018using}). These methods are reliable and easy to interpret, however, cannot handle complex scenarios. Thus, the learning-based method starts to draw the community's attention. In the static unstructured road, Fan, et al  \cite{fan2021learning} proposed to learn the conditional value-at-risk to represent the robot risk. However, a traffic scenario has more than one dynamic agent, which requires the robot to understand the interaction between multiple agents. In addition, relying solely on current information for risk quantification fails to accurately capture changes in the environment. It prevents the ego vehicle from making proactive reactions to dynamic conditions. Hence, the prediction-risk pipeline is necessary, however, the existing methods \cite{shao2024uncertainty, xu2023risk} just try to capture complex interaction with a simple model and ignore the occlusion issue and perception errors.

\noindent \textbf{V2X Perception and Prediction.} Due to the occlusion and limited perception range of single vehicles, multi-agent cooperative systems with V2X have enabled a new paradigm \cite{yangjie2024towards}. The major V2X work focuses on cooperative perception \cite{xu2022v2x, xiang2023hm}, which shares single-frame perception information to see through the occlusion. Recently, the temporal cue in V2X information has also attracted research attention \cite{zhou2024v2xpnp}, which fuses temporal information to support the temporal task, such as cooperative prediction \cite{ruan2023learning, zhang2025co}, and the temporal information can also be complemented with V2X. Recent V2X datasets in simulation and real-world, e.g., OPV2V \cite{xu2022opv2v}, V2X-Real \cite{xiang2024v2x}, and V2XPnP \cite{zhou2024v2xpnp}, also accumulate the cooperation automation research. Nonetheless, there is no work to further explore how to employ the V2X information to have a comprehensive risk quantification.

\noindent \textbf{Trajectory Prediction.} Trajectory prediction has drawn the community’s attention \cite{huang2022survey}, and the features of interaction, map, and history information have been considered in the emerging model \cite{zhao2021tnt, shi2022motion}, with the promising deep learning method. However, those prediction methods focus on the single agent prediction \cite{ngiam2021scene}, and the scenario behavior is described by each agent's isolated prediction behaviors. The prediction results from different objects will have self-collision and narrow the safe driving area for the following planning module. Risk-oriented prediction should be multi-modal and multi-agent. Transformer can be a typical model base for the multi-modal multi-agent decoder \cite{shi2024mtr++, zhou2023qcnext}, which can capture the trajectory feature in different temporal of different objects with self-attention.

\begin{figure*}[!t]
  \centering
  \includegraphics[width=\linewidth]{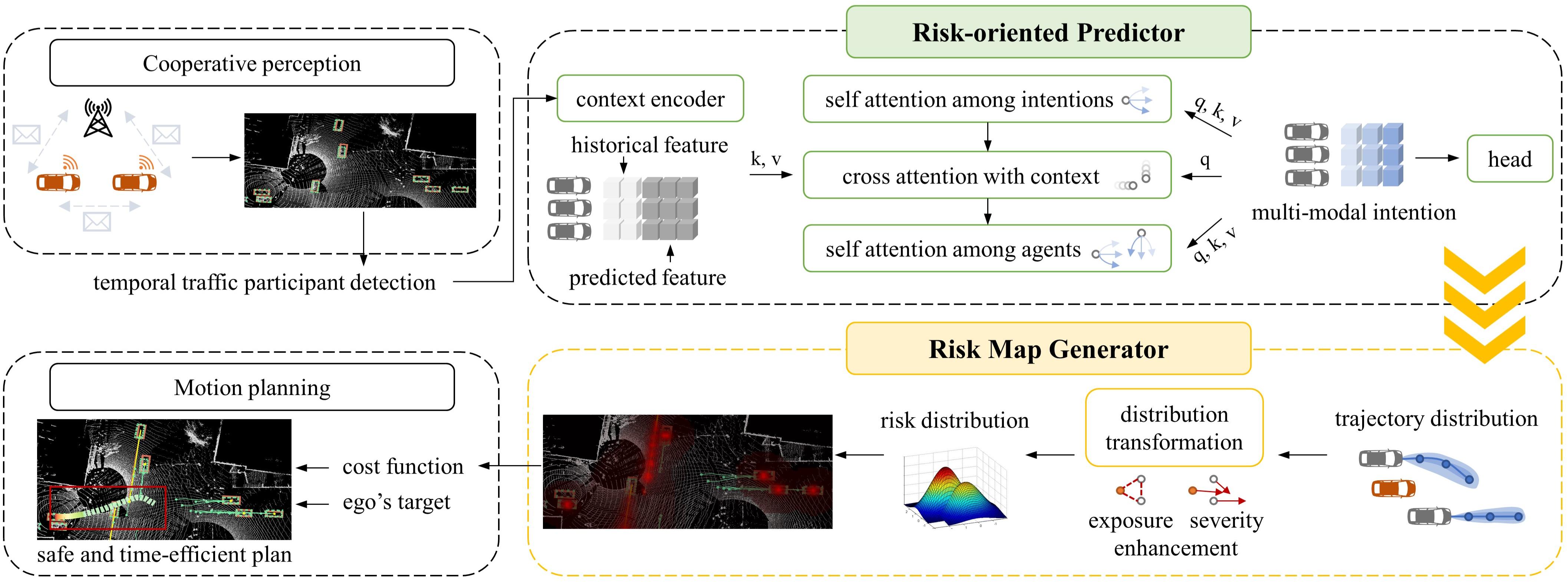}
  \caption{Overview of our \textit{CooperRisk} framework. (i) Cooperative perception: fuse multi-agent information and operate detection at each frame. (ii) Risk-oriented Predictor: incorporate intention-wise, context-wise and agent-wise attention to decode multi-modal trajectory distributions for multiple agents. (iii) Risk Map Generator: extract risk exposure and severity distributions from trajectory distributions to generate risk maps. (iv) Motion planning: an application module of \textit{CooperRisk}, utilizing risk maps as cost objectives to plan safe and time-efficient trajectories for the ego CAV.}
  \label{Figure2}
  \vspace{-0.6cm}
\end{figure*}

\section{METHODOLOGY}

Our proposed \textit{CooperRisk} pipeline comprises three main modules, as illustrated in Fig. \ref{Figure2}: (i)  Cooperative perception, (ii) Risk-oriented predictor, and (iii) Risk map generator. With the future states reasoning from the trajectory predictor, a risk map generator can provide temporal risk maps. Here, the risk map is defined as a map presenting the time, location, and risk value for the ego CAV. These maps function as the cost objectives for downstream motion planning.
% This method not only preserves necessary scene predictions, but also highlights the features of risk severity and risk exposure.

% This section presents the proposed CooperRisk pipeline for risk quantification and the method details.

% The proposed risk quantifier is established following the framework.

\subsection{Cooperative Perception}
To ensure the generalization to different agent types and sensors, while reducing the communication load in the multiple history frames, our \textit{CooperRisk} framework adopts the late fusion strategy, follow the real-world setting in \cite{zheng2024cooperfuse}.

\noindent \textbf{Metadata Sharing}. In a V2X scenario, each agent (CAV or infrastructure) collaborates with other agents within its communication range. V2X metadata (e.g. pose and extrinsic parameters) are shared first for establishing a spatial coordination among agents.

\noindent \textbf{Feature Extraction and Detection.} Each agent detects its surrounding traffic participants. Perception feature is extracted at each agent. We utilize the PointPillar network \cite{lang2019pointpillars} as the LiDAR feature backbone as the is extracted, known for its low inference latency. This process is executed at each timestamp and generates features in a sequence.

\noindent \textbf{Multi-agent Fusion.}The ego agent communicates with other agents and fuses the detected results from multiple agents at each historical frame via non-maximum suppression. Each agent in the V2X graph consolidates the shared information within a unified ego CAV coordinate system. In summary, the fused results capture the comprehensive scene information from multiple agents and frames.

\subsection{Scene-consistent Trajectory Prediction}
This part focuses on learning the scenario dynamics from the V2X perception results. The future motions are various and involve multi-modalities. A riskQ pipeline without differentiating the multi-modalities will assign the risk everywhere and generate a narrow driving area for the planning module. Thus, we propose a transformer-based module for multi-modal multi-agent prediction. It includes a scene context encoder and a motion decoder.

\noindent \textbf{Scene Context Encoder.} The proposed context encoder extracts features from the fused historical perception information. The information includes traffic participants' extents, locations, velocities, accelerations and heading angles, all tracked in a temporal sequence. The encoder utilizes a PointNet-like \cite{qi2017pointnet} polyline encoder to generate features of traffic participants as polyline vectors and utilizes a Transformer encoder to encode the historical dynamics of each traffic participant. Based on this encoding, the future dynamics of these traffic participants are predicted through a dense regression process. The future dynamics features are then combined with the historical dynamics features. The final encoding of the scene context is denoted as $\boldsymbol{F}_{c} \in \mathbb{R}^{N \times D}$, where $N$ is the number of traffic participants in the scene and $D$ is the feature dimension. 
It captures both historical and future motion features of each traffic participant.

\noindent \textbf{Multi-modal Multi-agent Motion Decoder.} The proposed motion decoder transforms the scene context feature into trajectory distributions. It utilizes a Transformer decoder to model the interactions among traffic participants. Considering the multi-modal nature of driving behaviors, each intention of a traffic participant corresponds to a query. Scene consistency is ensured by pairing the intentions of any two traffic participants. By guaranteeing that the predicted intentions are mutually consistent, this module produces more realistic and scene-consistent trajectory predictions. According to this strategy, agent-level self-attention, cross-attention and scene-level self-attention are implemented sequentially. Agent-level self-attention queries the interaction between different intentions of a traffic participant. Cross-attention queries the influence of the scene context on these intentions. Finally, scene-level self-attention queries the interaction between different traffic participants within the same scene. The process is updated by multi-head self-attention (MHSA) and multi-head cross-attention (MHCA) as follows:
\vspace{-0.5em}
\begin{equation}
    \boldsymbol{F}_{1}=M H S A\left(Q:\left[\boldsymbol{I}, \boldsymbol{P}_{I}\right], K:\left[\boldsymbol{I}, \boldsymbol{P}_{I}\right], V: \boldsymbol{I}\right)
    \label{Equation1}
\end{equation}
\begin{equation}
    \boldsymbol{F}_{2}=M H C A\left(Q:\left[\boldsymbol{F}_{1}, \boldsymbol{P}_{I}\right], K:\left[\boldsymbol{F}_{c}, \boldsymbol{P}_{c}\right], V: \boldsymbol{F}_{c}\right)
    \label{Equation2}
\end{equation}
\begin{equation}    \boldsymbol{F}_{3}=\operatorname{MHSA}\left(Q:\left[\boldsymbol{F}_{2}{ }^{\prime}, \boldsymbol{P}_{I}{ }^{\prime}\right], K:\left[\boldsymbol{F}_{2}{ }^{\prime}, \boldsymbol{P}_{I}{ }^{\prime}\right], V: \boldsymbol{F}_{2}{ }^{\prime}\right)
    \label{Equation3}
\end{equation}
where $\boldsymbol{I} \in \mathbb{R}^{N \times M \times D}$ is the intention features, $M$ is the number of modes (the number of intentions of a traffic participant), $\boldsymbol{P}_{I} \in \mathbb{R}^{N \times M \times D}$ is the position embedding of $\boldsymbol{I}$, $\boldsymbol{P}_{c} \in \mathbb{R}^{N \times D}$ is the position embedding of $\boldsymbol{F}_{c}$. Here, $\boldsymbol{F}_{c}$ and $\boldsymbol{P}_{c}$ should be expanded to $\mathbb{R}^{N \times N \times D}$ to match the query in equation \eqref{Equation2}. $\boldsymbol{F}_{2}{ }^{\prime} \in \mathbb{R}^{M \times N \times D}$ is derived by permuting the first and second dimensions of $\boldsymbol{F}_{2}$.

\noindent These informed queries are then decoded using a Gaussian Mixture Model (GMM), with the decoding result following Gaussian distributions. The output trajectory distributions are denoted as $\boldsymbol{Z} \in \mathbb{R}^{M \times N \times K \times E}$, where $K$ is the prediction horizon, and $E$ is the feature dimension, including means $(\mu_{x}, \mu_{y})$, variances $(\sigma_{x}, \sigma_{y})$ and covariance $\boldsymbol{\Sigma}$. The weighted scores for each mode are also provided as $\boldsymbol{W} \in \mathbb{R}^{M \times N}$.

\subsection{Risk Map Generation}
This part contributes to identifying potential risks involving the ego CAV in the scenario. The risk arises from the interaction between the ego CAV and the background traffic participants. As the above part has learned the scene dynamics in the V2X scenario, we introduce a distribution transformation to convert the trajectory distributions of traffic participants into a risk distribution of ego CAV based on risk severity and exposure.

Firstly, the risk measurement of the interaction between two objects is formulated as follows:
\vspace{-0.5em}
\begin{equation}
    V=\frac{c_{0} \Delta v^{2}+c_{1}}{\sqrt{c_{2}\left(\frac{\Delta s}{e^{c_{3} \Delta v}}\right)^{2}+c_{4} \Delta l^{2}}}
    \label{Equation4}
\end{equation}
risk severity indicator:
\vspace{-0.5em}
\begin{equation}
    \Delta v=\textstyle\frac{m_{A}}{m_{A}+m_{B}} \sqrt{v_{A}^{2}+v_{B}^{2}-2 v_{A} v_{B} \cos \alpha}
    \label{Equation5}
\end{equation}
risk exposure indicator:
\vspace{-0.5em}
\begin{equation}
    \Delta s=s^{*}-s_{0}^{*}
    \label{Equation6}
\end{equation}
\vspace{-1.5em}
\begin{equation}
    \Delta l=l^{*}-l_{0}^{*}
    \label{Equation7}
\end{equation}
\vspace{-1.3em}
\begin{equation}
    \left[\begin{array}{l}
    s^{*} \\
    l^{*}
    \end{array}\right]=\left[\begin{array}{cc}
    \cos \varphi & \sin \varphi \\
    -\sin \varphi & \cos \varphi
    \end{array}\right]\left[\begin{array}{l}
    s \\
    l
    \end{array}\right]
    \label{Equation8}
\end{equation}
\vspace{-0.7em}
\begin{equation}
    \left[\begin{array}{l}
    s_{0}^{*} \\
    l_{0}^{*}
    \end{array}\right]=\left[\begin{array}{cc}
    \cos \varphi_{0} & \sin \varphi_{0} \\
    -\sin \varphi_{0} & \cos \varphi_{0}
    \end{array}\right]\left[\begin{array}{l}
    s_{0} \\
    l_{0}
    \end{array}\right]
    \label{Equation9}
\end{equation}
where $V$ is the risk value, $m_{A}$, $v_{A}$, $s$, $l$, $\varphi$ represent the mass, speed, longitudinal position, lateral position and heading angle of the ego CAV, $m_{B}$, $v_{B}$, $s_{0}$, $l_{0}$, $\varphi_{0}$ represent the mass, speed, longitudinal position, lateral position and heading angle of the traffic participant. $\alpha$ is the relative angle between the two speed vectors. $s^{*}$, $l^{*}$, $s_{0}^{*}$ and $l_{0}^{*}$ are pseudo values of $s$, $l$, $s_{0}$ and $l_{0}$, considering the impact of the heading angle on the conflict possibility between two objects. $c_{0}$, $c_{1}$, $c_{2}$, $c_{3}$ and $c_{4}$ are parameters with positive values.

\noindent \textbf{Risk Severity} can be measured by introducing $\Delta v$, which is defined as the relative speed related to a conflict. The formulation is motivated by the fact that the masses and speeds of the two objects influence the severity of the risk. 

\noindent \textbf{Risk Exposure} can be measured by introducing $\frac{1}{\sqrt{\Delta s^{2}+\Delta l^{2}}}$. When the ego CAV gets closer to a traffic participant, the risk exposure increases. Considering that $\Delta v$ affects the rate at which the ego CAV approaches the traffic participant, $\Delta v$ is adopted as a scaling factor in equation \eqref{Equation4}.

\noindent \textbf{Risk Distribution.} When treating the location of the ego CAV as the independent variables and the location of the traffic participant as the parameters, the above risk measurement becomes a distribution $V\left(s, l \mid s_{0}, l_{0}\right)$. According to the Section III.B, $s_{0}$ and $l_{0}$ follows a Gaussian distribution $G\left(s_{0}, l_{0} \mid \mu_{s}, \mu_{l}, \sigma_{s}, \sigma_{l}, \boldsymbol{\Sigma}\right)$. Hence, the risk value follows a hierarchical distribution $V^{h}\left(s, l \mid \mu_{s}, \mu_{l}, \sigma_{s}, \sigma_{l}, \boldsymbol{\Sigma}\right)$. At each timestamp in the prediction horizon, the modes of each traffic participant are accumulated with their corresponding weighted scores, resulting in the risk distribution as follows:
\vspace{-0.7em}
\begin{equation}
    \begin{array}{l}
    V^{h}\left(s, l, t \mid \mu_{s, i, j, t}, \mu_{l, i, j, t}, \sigma_{s, i, j, t}, \sigma_{l, i, j, t}, \boldsymbol{\Sigma}_{i, j, t},  w_{i, j, t}\right),\\
    i \in[1, N], j \in[1, M], t \in[0, K]
    \end{array}
    \label{Equation10}
\end{equation}

\noindent \textbf{Risk Map.} By mapping the risk distribution across the entire scene, a risk map is derived. The risk value is the quantity of the distribution corresponding to different variables, namely the time and location where the risk occurs.

\subsection{Risk Map Based Planning}

% This section demonstrates how the proposed risk map enables downstream planning tasks. 
Optimization-based planning methods have been the mainstream approach for CAVs. Model predictive control (MPC) is adopted in this paper to leverage the above risk predictions in planning. The system state $\boldsymbol{X}$ and the control $\boldsymbol{U}$ are:
\vspace{-0.5em}
\begin{equation}
    \boldsymbol{X}_{k}=\left[\begin{array}{llll}
    s^{p}_{k} & v^{p}_{k} & l^{p}_{k} & \varphi^{p}_{k}
    \end{array}\right]^{T}, k \in[0, K]
    \label{Equation11}
\end{equation}
\vspace{-1.5em}
\begin{equation}
    \boldsymbol{U}_{k}=\left[\begin{array}{ll}
    a_{k} & \delta_{k}
    \end{array}\right]^{T}, k \in[0, K-1]
    \label{Equation12}
\end{equation}
where $s^{p}_k$, $v^{p}_k$, $l^{p}_k$, $\varphi^{p}_k$, $a_k$ and $\delta_{k}$ represent the longitudinal position, speed, lateral position, heading angle, acceleration and front wheel angle of the ego CAV at control step $k$, and $K$ is the control horizon. 

Vehicle dynamics model is formulated as follows.
\vspace{-0.5em}
\begin{equation}
    \boldsymbol{X}_{k+1}=\boldsymbol{A}_{k} \boldsymbol{X}_{k}+\boldsymbol{B}_{k} \boldsymbol{U}_{k}, k \in[0, K-1]
    \label{Equation13}
\end{equation}
\vspace{-1.4em}
\begin{equation}
    \boldsymbol{A}_{k}=\Delta t \times\left[\begin{array}{cccc}
    0 & 1 & 0 & 0 \\
    0 & 0 & 0 & 0 \\
    0 & 0 & 0 & v^{p}_{k} \\
    0 & 0 & 0 & 0
    \end{array}\right]+\boldsymbol{I}_{4 \times 4}, k \in[0, K-1]
    \label{Equation14}
\end{equation}
\vspace{-0.5em}
\renewcommand{\arraystretch}{0.5}
\begin{equation}
    \boldsymbol{B}_{k}=\Delta t \times\left[\begin{array}{cc}
    0 & 0 \\
    1 & 0 \\
    0 & 0 \\
    0 & \frac{v^{p}_{k}}{l_{f r}}
    \end{array}\right], k \in[0, K-1]
    \label{Equation15}
\end{equation}
where $\Delta t$ is the time increment in each control step and $l_{f r}$ is the distance between the ego front axle and rear axle.

\begin{table*}[t]
\caption{Detection, prediction, risk-map-based planning comparison of \textit{CooperRisk} in the real-world V2XPnP Dataset.}
\label{Table1}
\centering
\setlength{\tabcolsep}{2.3mm}
\renewcommand{\arraystretch}{1.23}
\begin{tabular}{ccc|c|cc|c|cc} % Increase column width
\toprule[1.1pt]
Perception & Prediction & Planning & AP@0.5 (\%)$\uparrow$ & minADE (m)$\downarrow$ & minFDE (m)$\downarrow$ & EPA$\uparrow$ & TOR$\downarrow$ & CR$\downarrow$ \\ \midrule 
\multirow{5}{*}{V2X sharing} & CV & \multirow{5}{*}{\shortstack{MPC planner\\based on risk map}} & 58.0 & 2.10 & 4.00 & 0.32 & / & / \\
& LSTM & & 58.0 & 1.20 & 2.25 & 0.40 & / & / \\ 
& QCNet \cite{zhou2023query} & & 58.0 & 0.80 & 1.40 & 0.47 & 0.07 & 0.08 \\ 
& MTR \cite{shi2022motion} & & 58.0 & 0.77 & 1.30 & 0.48 & 0.09 & 0.08 \\ 
\rowcolor{gray!20} & \textbf{CooperRisk} & & 58.0 & \textbf{0.74} & \textbf{1.26} & \textbf{0.50} & \textbf{0.04} & \textbf{0.04} \\ 
\midrule 
\multirow{5}{*}{Single-agent} & CV & \multirow{5}{*}{\shortstack{MPC planner\\based on risk map}} & 51.0 & 2.37 & 4.57 & 0.28 & / & / \\
& LSTM & & 51.0 & 1.40 & 2.59 & 0.34 & / & / \\ 
& QCNet \cite{zhou2023query} & & 51.0 & 1.32 & 2.37 & 0.35 & 0.06 & 0.09 \\ 
& MTR \cite{shi2022motion} & & 51.0 & 1.10 & 1.71 & 0.36 & 0.05 & 0.09 \\ 
\rowcolor{gray!20} & \textbf{CooperRisk} (no fusion) & & 51.0 & \textbf{1.06} & \textbf{1.65} & \textbf{0.37} & \textbf{0.04} & \textbf{0.08} \\ 
\midrule 
\multirow{5}{*}{Ground truth} & CV & \multirow{5}{*}{\shortstack{MPC planner\\based on risk map}} & / & 2.05 & 3.98 & / & / & / \\ 
& LSTM & & / & 1.02 & 1.99 & / & / & / \\ 
& QCNet \cite{zhou2023query} & & / & 0.75 & 1.29 & / & 0.08 & 0.00 \\ 
& MTR \cite{shi2022motion} & & / & 0.58 & 0.97 & / & 0.11 & 0.00 \\ 
\rowcolor{gray!20} & \textbf{CooperRisk} (ground truth) & & / & \textbf{0.56} & \textbf{0.94} & / & \textbf{0.07} & \textbf{0.00} \\ 
\bottomrule[1.1pt]
\end{tabular}%
\vspace{-0.3cm}
\end{table*}

The objective of the planning task is to enhance driving safety while maintaining mobility for the ego CAV. Adopting the proposed risk map $V^{h}$, the cost function is as follows:
\begin{equation}
    J=\min \smash{\sum_{k=0}^{K-1}\left[V^{h}\left(\boldsymbol{X}_{k}\right)+\left(\boldsymbol{X}_{k}-\boldsymbol{X}_{k}^{d}\right)^{T} \boldsymbol{Q}\left(\boldsymbol{X}_{k}-\boldsymbol{X}_{k}^{d}\right)\right]}
    \label{Equation16}
\end{equation}
\begin{equation}
    \boldsymbol{X}_{k}^{d}=\left[s_{k}^{p,*}, v^{*}, 0,0\right]^{T}, k \in[0, K-1]
    \label{Equation17}
\end{equation}
where $\boldsymbol{X}_{k}^{d}$ is the desired system state vector at control step $k$, $v^{*}$ is the ego CAV's expected speed, $s_{k}^{p,*}$ is the ego CAV's desired longitudinal position when driving with $v^{*}$ at control step $k$, and $\boldsymbol{Q}$ is a weighting matrix. Minimizing $V^{h}\left(\boldsymbol{X}\right)$ contributes to reduced risk and minimizing $(\boldsymbol{X}-\boldsymbol{X}^{d})^{T} \boldsymbol{Q}\left(\boldsymbol{X}-\boldsymbol{X}^{d}\right)$ contributes to enhanced mobility.

Error backpropagation is adopted to solve the nonlinear optimal control problem. Several iterations are performed to update $\boldsymbol{U}$ and $\boldsymbol{X}$. For each control step $k$ at each iteration $i$, $\boldsymbol{U}$ and $\boldsymbol{X}$ are updated as follows:
\vspace{-0.5em}
\begin{equation}
    \boldsymbol{U}_{k}^{i+1}=\boldsymbol{U}_{k}^{i}-\nabla_{\boldsymbol{U}} J\left(\boldsymbol{U}_{k}^{i}\right), k \in[0, K-1], i \in[0, I-1]
    \label{Equation18}
\end{equation}
\begin{equation}
    \begin{array}{l}
    \nabla_{U} J\left(\boldsymbol{U}_{k}^{i}\right)=\left[\frac{\partial J}{\partial \boldsymbol{U}}\right]_{\boldsymbol{U}_{k}^{i}}^{T} J\left(\boldsymbol{U}_{k}^{i}\right) \\
    =\left[\boldsymbol{B}_{k}^{i}\right]^{T} \frac{\partial V^{h}}{\partial \boldsymbol{X}} V^{h}\left(\boldsymbol{X}_{k}^{i}\right) \\
    +\left[\boldsymbol{B}_{k}^{i}\right]^{T} \boldsymbol{Q}\left(\boldsymbol{X}_{k}^{i}-\boldsymbol{X}_{k}^{d, i}\right)\left(\boldsymbol{X}_{k}^{i}-\boldsymbol{X}_{k}^{d, i}\right)^{T} \boldsymbol{Q}\left(\boldsymbol{X}_{k}^{i}-\boldsymbol{X}_{k}^{d, i}\right)
    \end{array}
    \label{Equation19}
\end{equation}
\begin{equation}
    \boldsymbol{X}_{k+1}^{i+1}=\boldsymbol{A}_{k}^{i+1} \boldsymbol{X}_{k}^{i+1}+\boldsymbol{B}_{k}^{i+1} \boldsymbol{U}_{k}^{i+1}, k \in[0, K-1], i \in[0, I-1]
    \label{Equation20}
\end{equation}
where $\boldsymbol{X}_{k}^{i}$ is the system state at control step $k$ during iteration $i$, and $I$ is the max number of iterations. The ego CAV's planned trajectory is obtained from $\boldsymbol{X}_{k}^{I}, k \in[0, K]$.

\begin{figure*}[h]
  \centering
  \includegraphics[scale=0.235]{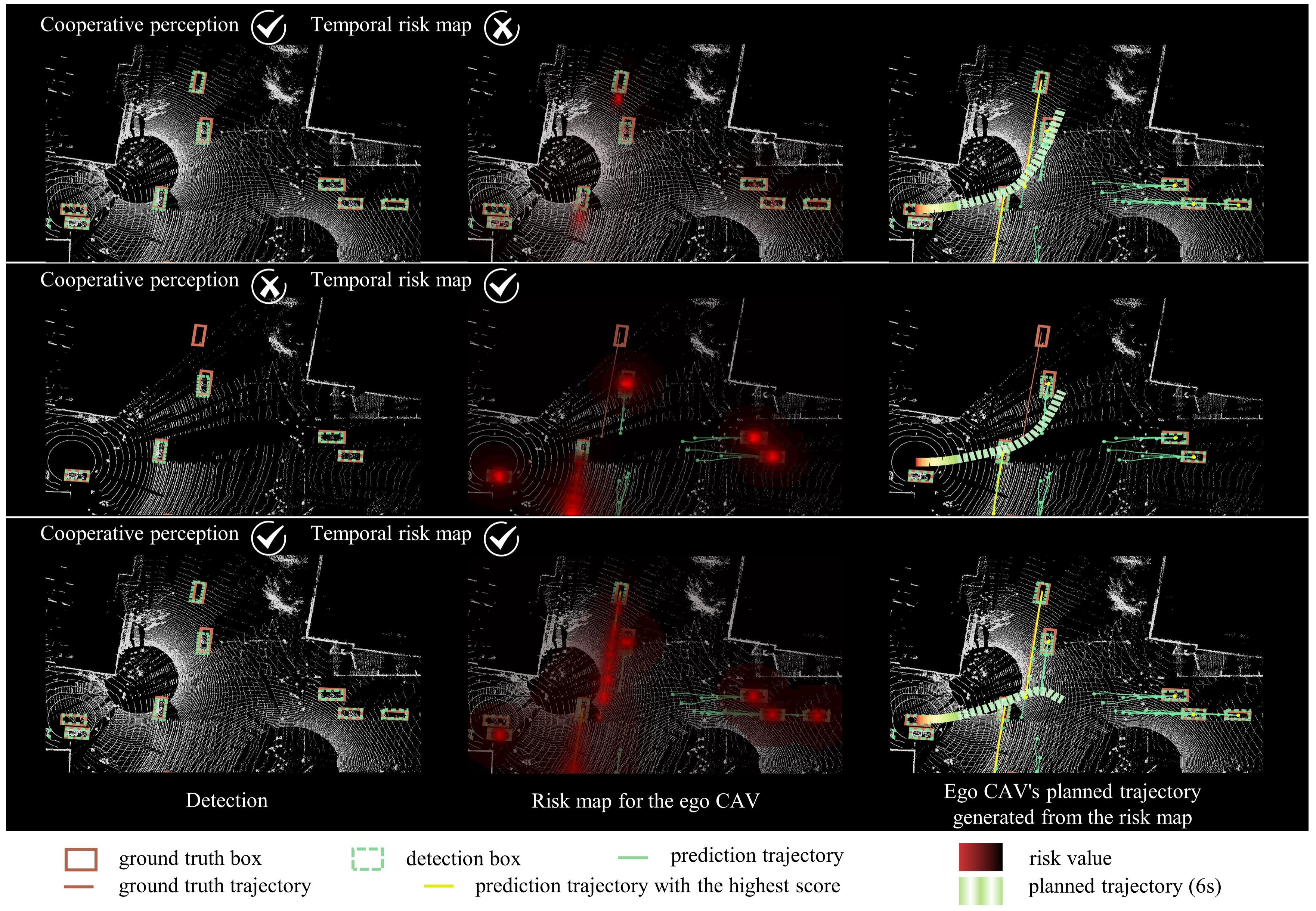}
  \caption{Qualitative results of \textit{CooperRisk} in the real-world V2XPnP Dataset. This scenario is chosen due to its dense traffic and complex interactions among objects, which provide a challenging and realistic environment for evaluating the performance of \textit{CooperRisk}. Each row represents different perception and risk map settings. The first column displays the detection results, the second column illustrates the generated risk maps, and the third column shows the ego CAV's planned trajectories derived from the risk maps.}
  \label{Figure3}
  \vspace{-0.4cm}
\end{figure*}

\section{EXPERIMENT}

\subsection{Evaluation Metrics}

The goal of the evaluation is to measure the accuracy of scenario awareness and the effectiveness of risk quantification in planning. Hence, Average Precision (AP), Minimum Average Displacement Error (minADE), Minimum Final Displacement Error (minFDE), Trajectory Overlap Rate (TOR), End-to-end Perception and Prediction Accuracy (EPA) \cite{gu2023vip3d} and Collision Rate (CR) are adopted as evaluation metrics. AP is used to evaluate the detection performance of the proposed pipeline with an intersection over the union threshold of 0.5. minADE, minFDE and TOR are used to evaluate the prediction performance of the proposed pipeline, with minADE and minFDE focusing on accuracy, and TOR emphasizing scene compliance. TOR is defined as the average probability of overlap in 3-dimension space between one object's predicted trajectories and other objects' predicted trajectories. EPA is used to evaluate the performance of joint perception and prediction.
\begin{equation}
    EPA=\smash{\frac{\left|\widehat{N}_{T P}\right|-\alpha N_{F P}}{N_{G T}}}
    \label{Equation21}
\end{equation}
where $|\widehat{N}_{T P}|$ is the number of true positive objects with prediction $minF D E<\tau_{E P A}$, $\tau_{E P A}$ is a threshold of prediction accuracy, $N_{F P}$, $N_{G T}$ are the number of false positive objects and ground truth objects, respectively, and $\alpha$ is a coefficient. $\tau_{E P A}$ is set to 2$\mathrm{m}$ and $\alpha$ is set to 0.5, following the settings in \cite{gu2023vip3d}. CR is used to evaluate the performance of the proposed risk map based planning. CR is defined as the probability of collision between the ego CAV's planned trajectory and the ground truth trajectories of background traffic participants.

\subsection{Baseline Methods}

\noindent \textbf{Trajectory Prediction Method.} Constant Velocity (CV) predictor, Long Short-Term Memory (LSTM) predictor, Query Context Network (QCNet) predictor \cite{zhou2023query} and Motion TRansformer (MTR) predictor \cite{shi2022motion} are chosen as baseline trajectory prediction methods. CV and LSTM predictors produce single-modal trajectory results, while QCNet and MTR predictors generate multi-modal ones. We reimplement the baselines model, and replace the predictor with the aforementioned predictors in our framework while maintaining all other modules unchanged. The comparison is for evaluating the prediction module's performance within the pipeline and its effectiveness in facilitating downstream planning task.

\noindent \textbf{Perception Information Sharing Mode.} To validate the influence of cooperative perception, a pipeline with single-agent perception is implemented in comparison, which relies solely on the ego CAV's perception. Additionally, in order to evaluate the influence of perception errors on \textit{CooperRisk}, a pipeline that directly uses ground truth scenario information is also implemented for comparison.

\subsection{Dataset and Implementation Details}

\noindent \textbf{Dataset.} V2XPnP dataset \cite{zhou2024v2xpnp} is utilized for the training and evaluation. It is the first real-world sequential V2X dataset collected by multiple mixed agents: two CAVs and two infrastructures. It covers 24 urban intersections, including roundabouts, T-junctions, and crossroads, totaling 40k frames of perception data. The captured traffic is dense, with diverse interactive driving behaviors. Therefore, it is appropriate to select the V2XPnP dataset for the training and evaluation of the proposed risk quantification pipeline.

\noindent \textbf{Implementation Details.} Following the setting in \cite{zhou2024v2xpnp, zheng2024cooperfuse}, the evaluation range is $x \in[-70.4,70.4] \mathrm{m}$, $y \in[-40,40] \mathrm{m}$. The communication range is set to 50$m$. The historical horizon is 2$s$ (2Hz), and the future horizon for prediction is 3$s$ (2Hz). The optimizer uses Adam with a weight decay of 0.01, and the learning rate scheduler adopts LambdaLR with an initial learning rate of 0.0001 and a 0.5 decay factor.

\subsection{Results Analysis}

Table \ref{Table1} presents the performance of scenario awareness and risk-map-based planning. Qualitative results are demonstrated in Fig. \ref{Figure3}.

\noindent \textbf{Accuracy of Scenario Awareness}. Table \ref{Table1} presents a comparison of different trajectory prediction methods in the proposed pipeline, evaluated in terms of minADE and minFDE. It should be noted that, to ensure the consistency of perception impact, AP values of the evaluations under the same perception information sharing mode remain identical. It is presented that \textit{CooperRisk} outperforms others in terms of prediction accuracy, with average improvements of 3.9\%, 7.5\% and 38.33\% in minADE compared to the MTR, QCNet and LSTM based pipeline, respectively. The rationale is that our method effectively differentiates multi-modal intentions, thereby pushing the boundaries of prediction accuracy compared to single-modal approaches such as LSTM and CV predictors. Although MTR and QCNet predictors are also multi-modal, our method surpasses them by effectively considering the scene consistent of multi-agent prediction. This capability contributes to more reasonable and accurate predictions. Furthermore, \textit{CooperRisk} shows consistent benefit over various perception settings. The benefit in prediction accuracy contributes to the subsequent risk quantification.

\noindent \textbf{Risk Map and Safe Planning}. Based on the prediction results, the risk distribution is calculated and the risk-map-based planning is conducted. Table \ref{Table1} presents the performance of the proposed pipeline on CR. The comparison of CR is conducted exclusively among \textit{CooperRisk}, QCNet and MTR based pipelines to ensure fairness, since they generate multi-modal trajectories, whereas CV based and LSTM based pipelines produce single-modal trajectories. Regardless of the prediction methods employed, CR is kept under 9\%, validating the effectiveness of risk-map-based planning. Furthermore, \textit{CooperRisk} achieves the lowest CR. This is because accurate scenario awareness provides accurate risk awareness, and the risk map considers the severity and exposure based on the awareness and provides good guidance for planning and avoiding collision.

\noindent In the case illustrated in Fig. \ref{Figure3}, the ego CAV stops at the intersection stop line and determines its future motion. It leverages cooperative perception to detect oncoming vehicles from the left, perceives driving risks within the scenario and plans a future trajectory that navigates through the cross traffic. The planning horizon is set longer than the prediction horizon, to evaluate how the risk quantification influences the ego CAV's long-term maneuvers. For comparison, the planning performance under single-agent perception and the performance utilizing only the current timestep's risk map are demonstrated. For the test without cooperative perception, the ego CAV fails to detect all oncoming vehicles from the left, leading to a left-turn decision based on the assumption that an unprotected left turn can be safely executed. However, a left turn is more hazardous than proceeding straight in this scenario. For the test without future risk maps, the ego CAV overlooks the propagation of risks and also decides to turn left. These comparisons demonstrate that V2X information facilitates downstream risk quantification and planning, while also confirming the effectiveness of the involvement of prediction for risk quantification.

\noindent \textbf{Ablation Study on the Multi-agent Prediction}. Table \ref{Table1} presents a comparison in terms of TOR. Similar to the comparison of CR, the comparison of TOR is also conducted exclusively among \textit{CooperRisk}, QCNet and MTR based pipelines to ensure fairness, since they generate multi-modal trajectories. It is presented that \textit{CooperRisk} predicts more scene-consistent trajectories. It reduces TOR by 37.31\% on average. The superior performance can be attributed to our method's ability to consider interactions among objects while simultaneously predicting their future motions. By providing the future trajectories of all objects within the scenario in a coordinated manner, it avoids the occurrence of intersecting or conflicting future motions, rather than assigning the risk everywhere and narrowing the driving area.

\noindent \textbf{Ablation Study on V2X Information Sharing}. Table \ref{Table1} presents a comparison of different perception strategies in the proposed pipeline, evaluated in terms of AP. It is demonstrated that V2X sharing based pipeline outperforms single-agent based pipeline in terms of detection accuracy. Compared to \textit{CooperRisk} with no fusion, AP is increased by 12.07\% with detection results from V2X sharing. The improvement in detection accuracy is highly significant for the pipeline, as the adoption of ground truth detection leads to enhanced performance in downstream tasks, including prediction, risk quantification, and planning.

\begin{figure}[t]
  \centering
  \includegraphics[scale=0.3]{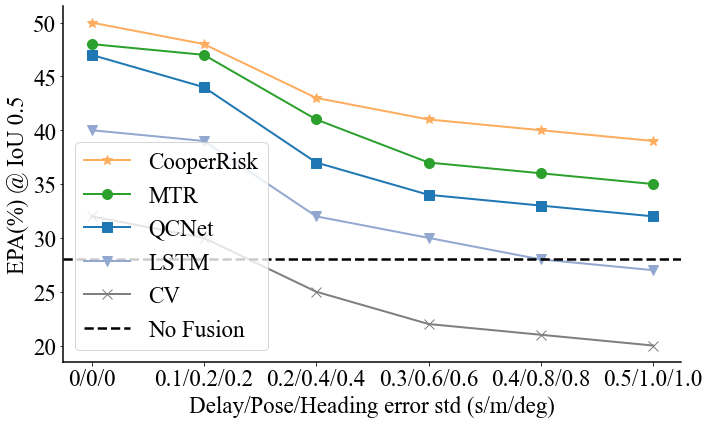}
  \caption{Communication noise and delay experiment}
  \label{Figure4}
  \vspace{-0.5cm}
\end{figure}

\noindent \textbf{Robustness Assessment on V2X Noise Settings}. Previous work always ignores the upstream module errors in riskQ and planning, such as perception and prediction. First, the aforementioned section (ablation study on V2X information sharing) shows the robustness of our pipeline with perception errors, especially the \textit{CooperRisk}. To further explore its robustness, Fig. \ref{Figure4} presents a comparison of pipelines based on various predictors under different V2X noise settings, evaluated in terms of EPA. The reason for selecting EPA is that: the performance of risk quantification is dependent on the effectiveness of prediction, which is also influenced by perception capability. EPA, which measures the joint accuracy of perception and prediction, serves as an appropriate metric for this evaluation. The positional and heading noises are drawn from a Gaussian distribution with standard deviations ranging from $0.2\mathrm{m}$ to $1.0\mathrm{m}$ for positional noise and from $0.2^{\circ}$ to $1.0^{\circ}$ for heading noise, respectively\cite{xu2022v2x, zhou2024v2xpnp}. The time delay is configured to range from $100\mathrm{ms}$ to $500\mathrm{ms}$. It is demonstrated that \textit{CooperRisk} achieves the best performance across all noise settings, exhibiting the least performance degradation with significant robustness.

\section{CONCLUSIONS}

In this paper, we introduce a cooperative perception-prediction-riskQ pipeline, \textit{CooperRisk}, which is the first driving-risk quantification pipeline with V2X information. We provide a multi-agent multi-modal predictor, ensuring that the trajectory predictions are scene-consistent for improved risk quantification. We present a new strategy for risk map generation and risk map-based planning, where the probability distribution-based risk map serves as an effective tool to bridge the prediction and planning tasks. The evaluation on a real-world V2X dataset V2XPnP shows the superior performance of \textit{CooperRisk} on the final task planning. The communication noise and delay experiment also show the strong robustness of \textit{CooperRisk}.

% \textbf{Limitation}. The proposed pipeline still needs to be validated with other V2X spatio-temporal fusion strategies, such as early fusion and intermediate fusion, to assess its practicality in a V2X environment. 

% Additionally, the proposed pipeline should be further trained and tested on datasets that include more collision scenarios.


%%%%%%%%%%%%%%%%%%%%%%%%%%%%%%%%%%%%%%%%%%%%%%%%%%%%%%%%%%%%%%%%%%%%%%%%%%%%%%%%

%%%%%%%%%%%%%%%%%%%%%%%%%%%%%%%%%%%%%%%%%%%%%%%%%%%%%%%%%%%%%%%%%%%%%%%%%%%%%%%%

%%%%%%%%%%%%%%%%%%%%%%%%%%%%%%%%%%%%%%%%%%%%%%%%%%%%%%%%%%%%%%%%%%%%%%%%%%%%%%%%

\section*{ACKNOWLEDGMENT}

This paper is partially supported by USDOT/FHWA Mobility Center of Excellence, National Science Foundation \# 2346267 POSE: Phase II: DriveX, National Science and Technology Major Project (No. 2022ZD0115501 or 2022ZD0115503), National Key R\&D Program of China (2022YFF0604905 or 2022YFE0117100), National Natural Science Foundation of China (Grant No. 52372317), Yangtze River Delta Science and Technology Innovation Joint Force (No. 2023CSJGG0800), Shanghai Automotive Industry Science and Technology Development Foundation (No. 2404), Xiaomi Young Talents Program, the Fundamental Research Funds for the Central Universities (22120230311).

\bibliographystyle{IEEEtran}
\bibliography{root}

\end{document}